\documentclass[10pt,twocolumn,letterpaper]{article}

\usepackage{wacv}
\usepackage{times}
\usepackage{epsfig}
\usepackage{array}
\usepackage{graphicx}
\usepackage{amsmath}
\usepackage{amssymb}
\usepackage{xcolor}
\usepackage{multirow}
\usepackage{amsfonts}
\usepackage{colortbl}
\usepackage{subcaption}
\usepackage{booktabs}
\usepackage{cite}

\newcommand{\wen}[1]{{\color{black}#1}} 

\def\wacvPaperID{302} 

\wacvfinalcopy 

\ifwacvfinal
\def\assignedStartPage{1} 
\fi

\ifwacvfinal
\usepackage[breaklinks=true,bookmarks=false]{hyperref}
\else
\usepackage[pagebackref=true,breaklinks=true,colorlinks,bookmarks=false]{hyperref}
\fi

\ifwacvfinal
\else
\fi

\newcommand{\pose}{\mathbf{p}}
\newcommand{\rpose}{\mathbf{q}}

\begin{document}

\title{PI-Net: Pose Interacting Network\\ for Multi-Person Monocular 3D Pose Estimation}

\author{Wen Guo$^1$, Enric Corona$^2$, Francesc Moreno-Noguer$^2$, Xavier Alameda-Pineda$^1$\\
$^1$Inria, Univ. Grenoble Alpes, CNRS, Grenoble INP, LJK, 38000 Grenoble, France\\
$^2$Institut de Robòtica i Informàtica Industrial, CSIC-UPC, Barcelona, Spain\\
$^1${\tt\small \{wen.guo,xavier.alameda-pineda\}@inria.fr}, $^2${\tt\small \{ecorona,fmoreno\}@iri.upc.edu}

}

\maketitle

\begin{abstract}
Recent literature addressed the monocular 3D pose estimation task very satisfactorily. In these studies, different persons are usually treated as independent pose instances to estimate. However, in many every-day situations, people are interacting, and the pose of an individual depends on the pose of his/her interactees. In this paper, we investigate how to exploit this dependency to enhance current -- and possibly future -- deep networks for 3D monocular pose estimation. Our pose interacting network, or PI-Net, inputs the initial pose estimates of a variable number of interactees into a recurrent architecture used to refine the pose of the person-of-interest. Evaluating such a method is challenging due to the limited availability of public annotated multi-person 3D human pose datasets. We demonstrate the effectiveness of our method in the MuPoTS dataset, setting the new state-of-the-art on it. Qualitative results on other multi-person datasets (for which 3D pose ground-truth is not available) showcase the proposed PI-Net. PI-Net is implemented in PyTorch and the code will be made available upon acceptance of the paper.
\end{abstract}
\vspace{2mm}
\section{Introduction}
\begin{figure*}
\begin{center}
\includegraphics[width=0.7\linewidth]{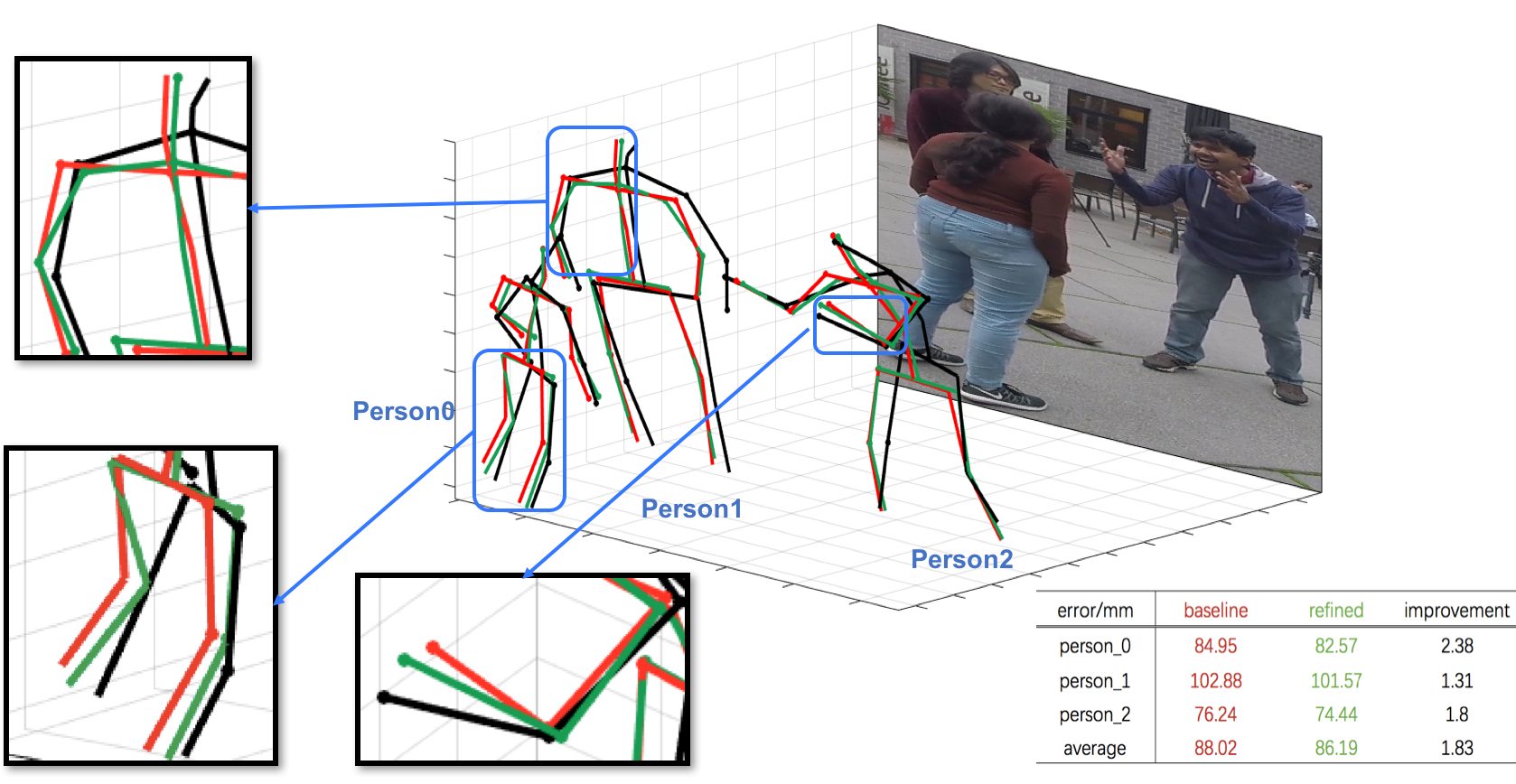}
  \caption{{\bf PI-Net peformance.} An example of testing on MuPoTS dataset. Poses refined by PI-Net (in green) are closer to the ground truth (in black) than the baseline (in red). We zoom-in to several parts to clearly appreciate the difference.  The error before and after PI-Net refinement for each person is shown in the table. The average 3D joint error for this example  is reduced from 88.02 mm to 86.19 mm.}
\label{fig:teaser}
\end{center}
\end{figure*}

Monocular 3D multi-person human pose estimation aims at estimating the 3D joints of several people from a single RGB image. This problem attracts great research and industrial interests, as it would make  possible a number of applications in many different fields including the entertainment industry, sports technology, physical therapy and medical diagnosis. 
Recent works on multi-person human pose estimation usually regard different people as independent instances and estimate the poses one by one in separate bounding boxes in top-down methods. This makes all these approaches agnostic about the context information and specifically about the presence of other people \cite{lcr,3dmppe,lcr++,singleshot,kumarapu2020animepose,von2018recovering,mehta2019xnect,benzine2019deep,dabral2019multi}.  However, when people interact, the pose and motion of every person is typically dependent and correlated  to the body posture of the people he/she is interacting with. 
\newline
While  context information has been shown to be useful   in   tasks such as object detection~\cite{divvala2009empirical,pato2020seeing,barnea2019exploring}, motion prediction\cite{Corona2019ContextawareHM} or affordance estimation~\cite{corona2020ganhand}, to the best of our knowledge, it has not been \wen{well developed} 
before in a body pose estimation. In this paper, we  investigate how  these dependencies can be used to boost the performance of off-the-shelf  architectures for 3D human pose estimation.
\newline
Concretely, we propose a pose interacting network, PI-Net, which is fed with the 3D pose of a person of interest and an arbitrary number of body poses from other people in the scene, all of them computed with a context agnostic pose detector. These poses are potentially noisy, both in their absolute position in space as in the specific representation of the body posture. PI-Net is built using a recurrent network with a self-attention module that encodes the contextual information. Since it is unclear how to rank the contextual information, that is the pose of other persons, regarding the potential impact on the pose refinement pipeline, we make the very straightforward assumption that the potential of a person to refine the pose of the person-of-interest, is inversely proportional to the square of the distance between them. 
\newline
We thoroughly evaluate our approach on the MuPoTS dataset~\cite{singleshot}, and using the initial detections of 3DMPPE~\cite{3dmppe}, the current best performing approach on this dataset. PI-Net exhibits consistent improvement of the pose estimates provided by 3DMPPE in all sequences of the dataset, becoming thus, the new state-of-the-art (see one example in Fig.~\ref{fig:teaser}). Interestingly, note that PI-Net can be used as drop-in replacement for any other architecture that estimates 3D human pose. Additionally, the size of the network we propose is relatively small ($3.41M$ training parameters, while the baseline model has $36.25M$ parameters), enabling efficient training and introducing a marginal computational cost at test. Testing on one Geforce1070, PI-net just cost 0.007s on refining one person while the baseline cost 0.038s for detecting one root-centered pose and also extra time on obtaining the bounding boxes and roots. Our method is lightweight and consistently improves the baseline.
\section{Related Work}
\subsection{3D Single-person pose estimation}
Deep learning methods for single-person 3D pose estimation follow two different strategies. On one hand, there are algorithms that directly learn the mapping from image features to 3D poses~\cite{martinez2017simple, li20143d, pavlakos2017coarse, qiu2019cross, ci2019optimizing}. For instance, \cite{li20143d} propose a joint model for body part detectors and pose regression. Pavlakos~\etal~\cite{pavlakos2017coarse} introduce a U-Net architecture to recover joint-wise 3D heatmaps. Sun~\etal~\cite{sun2017compositional} build a regression approach using a bone-based representation that enforces human pose structure. In~\cite{sun2018integral}, a differentiable soft-argmax operation is used for efficiently training a hourglass network.

Another line of work  focuses on recovering 3D human pose from 2D image features by using models that enforce consistency between 3D predicted poses and 2D observations~\cite{wei2009, bogo2016, moreno20173d}. 
For instance, Bogo~\etal~\cite{bogo2016} fit a human body parametric model by minimising the distance between the projection of the 3D estimation and the 2D predicted joints. 
Moreno-Noguer~\cite{moreno20173d} propose to infer 3D pose via distance matrix regression. Yang~\etal~\cite{yang20183d} use an adversarial approach to ensure that estimated poses are antropomorphic.

\begin{figure*}
\begin{center}
\vspace{0.2cm}
\includegraphics[trim={0.6cm 0cm 1.65cm 0cm},clip=true,width=0.9\linewidth]{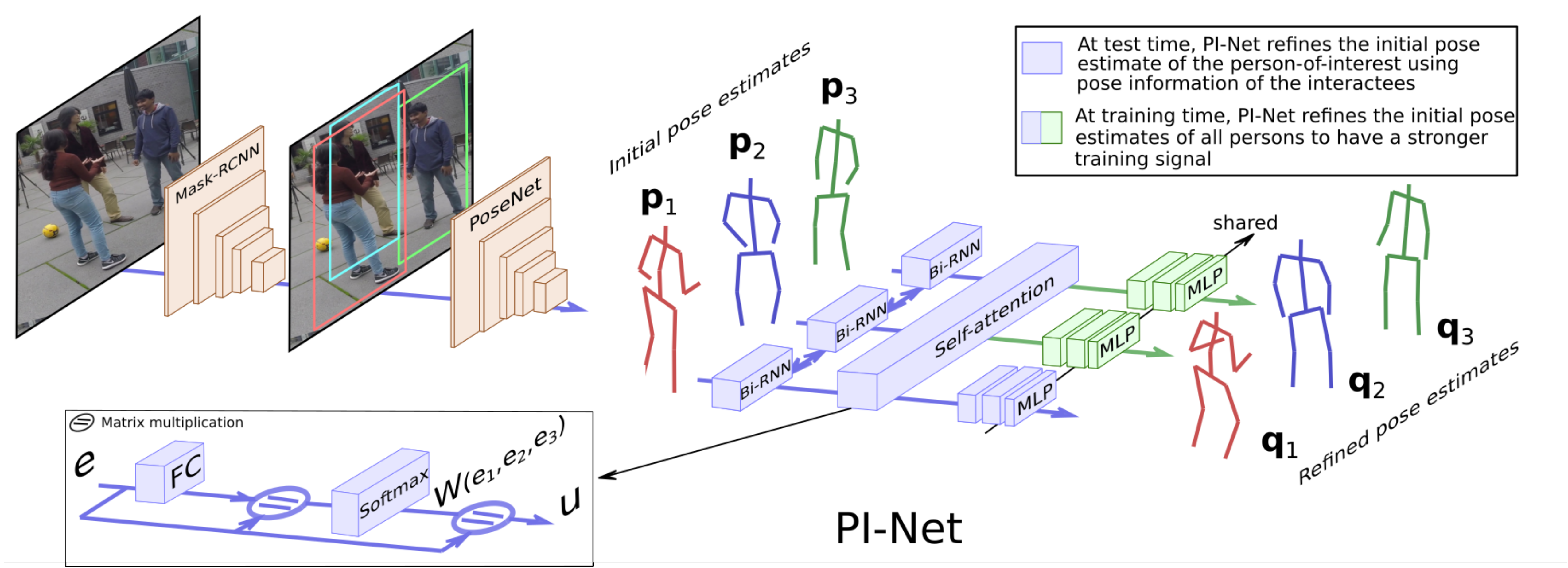}
\vspace{0cm}
  \caption{{\bf PI-Net Architecture}. Mask-RCNN~\cite{maskrcnn} and PoseNet~\cite{3dmppe} are used to extract the initial pose estimates $\pose_1,\ldots,\pose_N$. These estimates are fed into PI-Net, composed of three main blocks: Bi-RNN,  Self-attention and the shared fully-connected layers. The output of PI-Net refines the initial pose estimates by exploiting the pose of the interactees, yielding $\rpose_1,\ldots,\rpose_N$.
}
\label{fig:pipline}
\end{center}
\end{figure*}

\subsection{2D multi-person pose estimation}
There are two main approaches for multi-person pose estimation, top-down~\cite{xiao2018simple,li2019rethinking,sun2019deep,chen2018cascaded} and bottom-up models~\cite{cao2017realtime,cao2018openpose,pishchulin2016deepcut,newell2017associative}. On the former, a human detector first estimates the bounding boxes containing the person. Each detected   area is cropped and fed into the pose estimation network. The later also follows a two-stage pipeline, where a model first estimates all human body keypoints, and then groups them into each person using  clustering techniques. 

Cao~\etal~\cite{cao2017realtime,cao2018openpose} propose a real-time bottom-up method using Part Affinity Fields to group joints of different person. The efficiency of these bottom-up approaches makes them very appropriate to be used as a backbone for later lifting the 2D joints to 3D~\cite{dabral2019multi,pavlakos2019expressive}. 
The performance of bottom-up methods has been recently improved by to-down strategies. Xiao~\etal~\cite{xiao2018simple} use ResNet~\cite{he2016deep} as encoder and several deconvolutional layers as decoder to formulate a simple but effective baseline. Sun\etal~\cite{sun2019deep} connect the high-to-low resolution convolution streams in parallel to maintain richer semantic information. Chen \etal~\cite{chen2018cascaded} use a cascade pyramid network to refine the hard keypoints of the initial estimated results.

\subsection{3D Multi-person pose estimation \wen{and contextual information}}
Similar to their 2D counterparts, 3D multi-person poses estimation methods can be split into top-down~\cite{lcr, lcr++, 3dmppe, kumarapu2020animepose, weinzaepfel2020dope} and bottom-up\cite{singleshot,mehta2019xnect,zanfir2018deepintegrated} approaches. Mehta~\etal~\cite{singleshot,mehta2019xnect} follow a bottom-up strategy, by first estimating  three occlusion-robust location-maps~\cite{mehta2017vnect} and then modeling the association between body keypoints using Part Affinity Fields~\cite{cao2017realtime}. Zanfir~\etal~\cite{zanfir2018deepintegrated} formalize the problem of localizing and grouping people as a binary linear integer program and solve it by integrating a limb scoring model.

Rogez~\etal~\cite{lcr, lcr++}, in contrast, propose a top-down approach, where first, each person 2D bounding box is classified into one of the anchor clustered 3D poses. These poses are then refined in a coarse-to-fine manner.  
Moon~\etal~\cite{3dmppe} propose an architecture that simultaneously predicts the 3D absolute position of the root joint and reconstructs the relative 3D body pose of multiple people.
However, despite the fact that these works estimate the body pose of an arbitrary number of people, each person is processed using an independent pipeline that does not take into account the interactions between the rest of people or other contextual information.

\wen{
Recently, some works begin to pay attention to using contextual information in 3D pose estimation problem by integrating scene constraints\cite{zanfir2018sceneconst} or considering the depth-order to resolve the overlapping problem \cite{jiang2020depthordering, li2020hmor}. Jiang~\etal~\cite{jiang2020depthordering} propose a depth ordering-aware loss to consider the occlusion relationship and interpenetration of people in multi-person scenarios. Li~\etal~\cite{li2020hmor} divide human relations into different levels and define 3 corresponding losses to tell if the orders of different people or different joints are correct or not.
Though contextual information is considered in these works, they do not really explore the interaction relations between different people in the same activity. 
More recently, Fieraru~\etal~\cite{fieraru2020CHI3D} proposed a new dataset of human interactions with several daily interaction scenarios and proposed a framework based on contact detection  over model surface regions, but this dataset is not released yet.
}

In this paper, we propose a method that can be used in combination with the current state-of-the-art model~\cite{3dmppe} and boost its performance by looking at the whole group of humans. The proposed model is flexible and can be stacked after any 3D pose estimation model, independently of it being top-down or bottom-up.

\section{PI-Net for Multi-Person Pose Estimation}

Our goal is to exploit the interaction information between $N$ people so as to improve the estimation of their pose. We assume the existence of an initial 3D pose estimate $\pose_n\in\mathbb{R}^{J\times 3}$ of person $n=1,\ldots,N$, where $J$ is the number of estimated joints, \eg\ obtained from 3DMPPE~\cite{3dmppe}. All the $N$ poses are in absolute camera coordinates.

Formally, our goal is to improve the initial pose estimates, taking into account the pose of other people:
\begin{equation}
[\rpose_1,\ldots,\rpose_N] =  \mathbf{\Pi} (\pose_1,\ldots,\pose_N),
\label{equ:refine}
\end{equation}
where $\rpose_n\in\mathbb{R}^{J\times 3}$ denotes the pose of person $n$ improved with the information of the poses of the interactees.

While the idea is very intuitive, the research question is how to design PI-Net (i.e.\ $\mathbf{\Pi}$) so that it satisfies the following desirable criteria. Firstly, it shall work in environments with different number of people $N$, and  not fixed to a particular scenario. Secondly, the interaction information can be efficiently exploited and learned using publicly available datasets. Finally, it has to be generic enough to work with \textit{any} 3D monocular multi-person pose estimator.

\subsection{Pipeline of PI-Net}

Naturally, the fact that the number of people $N$ is unknown in advance, points us towards the use of recurrent neural networks. Such RNN should input the poses estimated by a generic pose estimator, and embed the pose information into a representation learned specifically to take the cross-interactions into account. Without loss of generality, let us assume that the person-of-interest is $n=1$, and hence the pose to refine is $\pose_1$. We consider using a bi-directionnal RNN, whose first input is $\pose_1$, and then the rest of initial poses are provided in a given order (see below). Our intuition for using a Bi-RNN is the following. During the forward pass, and since the first input is $\pose_1$, the network can use the information in $\pose_1$ to extract the features of the other poses that will best refine $\pose_1$. In the backward pass, the network accumulates all this information back to $\pose_1$, obtaining:
\begin{equation}
    \mathbf{e}_1 =  \textrm{Bi-RNN}(\pose_1,\ldots,\pose_N).
\end{equation}

The learned embedding \wen{ $\mathbf{e}_1\in\mathbb{R}^{N\times E}$}is supposed to contain the crucial information from all other poses to refine the pose of the person-of-interest ($1$ in our example), but not only. Indeed, given that a priori we do not know which persons would be more helpful in refining the pose of interest, the computed embedding $\mathbf{e}_1$ could contain information that is not exploitable to refine the pose. In order to take this phenomenon into account, we soften the requirements of the Bi-RNN through the use of an attention mechanism as shown in Figure~\ref{fig:pipline} (bottom-left zoom). Such attention mechanism aims to improve each embedding by combining information from the embeddings of other persons. To do so, we compute a matrix of attention weights:
\begin{equation}
\mathbf{W}\in\mathbb{R}^{N\times N}, \qquad  \mathbf{W}_{nm} = \mathbf{e}_n^\top(\mathbf{A}_{\textsc{att}}\mathbf{e}_m+\mathbf{b}_{\textsc{att}})),
\end{equation}
that is then normalised with a row-wise soft-max operation. $\mathbf{A}_{\textsc{att}}$ and $\mathbf{b}_{\textsc{att}}$ are attention parameters to be learned. The self-attention weights $\mathbf{W}$ encoding the residual interaction not captured by the Bi-RNN are used to update the embedding vector $\mathbf{u}_1 = {\mathbf{W}\mathbf{e}_1}$. Finally, the updated embedding is feed-forwarded through a few fully connected layers, obtaining the final refined pose $\rpose_1$. 
While, at test time the self-attention and fully-connected layers are used only for the person-of-interest, at training time we found it is useful to apply these two operations to all poses, and back-propagate the loss associated to everyone. This strategy eases the training. The overall pipeline depicting of PI-Net is shown in Figure~\ref{fig:pipline}.

\subsection{Interaction Order}
\label{sec:interaction}
In the previous section we assumed that the order in which the initial pose estimates $\pose_n$ were presented to the Bi-RNN was given. Although there is no principled rule to define the ordering, there are some requirements. For a given person $n$, the sequence of poses presented to the network $\pose_{\rho_n(1)},\ldots,\pose_{\rho_n(N)}$ has two constraints: (i) each pose is presented only once and (ii) the first pose is the one to be refined, i.e.\ $\rho_n(1)=n$. Intuitively, the order should represent the relevance: the more useful $\pose_m$ is to refine $\pose_n$, the closer $\pose_m$ should be to $\pose_n$ in the input sequence, i.e. the smaller $\rho_n(m)$ should be. Because finding the optimal permutation is a complex combinatorial optimisation problem for which there is no ground-truth, we opt for assuming that the relevance is highly correlated to the physical proximity between interactees. Therefore, the closer person $m$ is to person $n$, the smaller should $\rho_n(m)$ be. With this rule we order the initial pose estimates to be fed to the Bi-RNN.

We also consider of using Graph Convolutional Network~\cite{kipf2016semi} to model the interaction between different person. Considering a pair of input persons, the node of the graph represents the coordinate of all the joints of these two people, and the adjacency matrix learned from the input represents interaction between these joints. This strategy does not provide any performance increase, the results will be discussed in Section~\ref{sec:ablation}.

\subsection{Network Architecture}

In order to build and train our PI-Net, we first extract the initial poses using~\cite{3dmppe}. In the baseline, Mask-RCNN is used to detect the people present in the image. After that, the keypoint detector is applied to each image to detect the root-based poses and then project them into absolute camera coordinates. This keypoint detector is based on ResNet50 and 3 addition deconvolutional layers, following~\cite{sun2018integral}. The set of keypoints for each person in camera coordinates $\pose_n$, is therefore obtained. Note that this regressor gives all $J$ person joints, despite of partial occlusions, the corresponding occluded joints are hallucinated.

These initial pose estimates are then normalised with their mean and standard deviation, thus obtaining the input pose estimates of our PI-Net, $\{\pose_1,\ldots,\pose_N\}$. For each person $n$, we feed the PI-Net with the sequence of poses in the order appropriate for person $n$ (see Section~\ref{sec:interaction}). The output $\rpose_n$ of PI-Net is the refined pose for person $n$. PI-Net is trained with the $L_1$ loss between the refined poses and the ground-truth in 3D camera coordinates, added for all detected persons in the training image.

The Bi-RNN is implemented using three layers of gated recurrent units (GRU~\cite{cho2014gru}). The the self-attention layer provides a straightforward way to account for person pose interactions. After applying attention, the updated embedding goes through three fully connected layers to output the refined 3D pose in camera coordinates. These three fully connected layers are shared by all $N$ poses.
Consequently, the proposed PI-Net can be trained and evaluated using images with different number of people.

\begin{table*}[htbp!]
\caption{Sequence-wise 3DPCK comparison with state-of-the-art methods on the MuPoTS-3D dataset. The first three methods show the reported results in the corresponding paper, the fourth method and our model is tested with ground truth bounding boxes and roots. Higher value means better performance.\vspace{-0.5cm}}
\label{tab:sequencewise pck}
\begin{center}
\resizebox{\textwidth}{!}{\setlength{\tabcolsep}{0.7mm}{
\begin{tabular}{l c c c c c c c c c c c c c c c c c c c c c}
\toprule
Sequence & S1 & S2 & S3 & S4 & S5 & S6 & S7 & S8 & S9 & S10 & S11 & S12 & S13 & S14 & S15 & S16 & S17 & S18 & S19 & S20 & AVG\\
\midrule\midrule
\multicolumn{22}{l}{\textbf{Accuracy for all ground truths}} \\
LCR\cite{lcr} & 67.7 & 49.8 & 53.4 & 59.1 & 67.5 & 22.8 & 43.7 & 49.9 & 31.1 & 78.1 & 50.2 & 51.0 & 51.6 & 49.3 & 56.2 & 66.5 & 65.2 & 62.9 & 66.1 & 59.1 & 53.8 \\
Singleshot\cite{singleshot} & 81.0 & 60.9 & 64.4 & 63.0 & 69.1 & 30.3 & 65.0 & 59.6 & 64.1 & 83.9 & 68.0 & 68.6 & 62.3 & 59.2 & 70.1 & 80.0 & 79.6 & 67.3 & 66.6 & 67.2 & 66.0 \\
Xnect\cite{mehta2019xnect} & 88.4 & 65.1 & 68.2 & 72.5 & 76.2 & 46.2 & 65.8 & 64.1 & 75.1 & 82.4 & 74.1 & 72.4 & 64.4 & 58.8 & 73.7 & 80.4 & 84.3 & 67.2 & 74.3 & 67.8 & 70.4 \\
LCR++\cite{lcr++} & 87.3 & 61.9 & 67.9 & 74.6 & 78.8 & 48.9 & 58.3 & 59.7 & 78.1 & 89.5 & 69.2 & 73.8 & 66.2 & 56.0 & 74.1 & 82.1 & 78.1 & 72.6 & 73.1 & 61.0 & 70.6\\
PandaNet\cite{benzine2020pandanet} & - & - & - & - & - & - & - & - & - & - & - & - & - & - & - & - & - & - & - & - & 72.0\\
3DMPPE\cite{3dmppe} & 93.2 & 75.6 & 80.3 & 81.5 & 84.6 & 75.3 & \textbf{84.5} & 69.3 & 90.1 & 92.0 & 81.0 & 81.0 & 73.4 & 73.5 & 81.8 & 89.6 & 88.4 & 84.3 & 74.5 & 70.6 & 81.2 \\
PI-Net (ours) & \textbf{93.5} & \textbf{77.4} & \textbf{82.0} & \textbf{82.9} & \textbf{87.2} & \textbf{75.9} & 84.0 & \textbf{71.5} & \textbf{90.2} & \textbf{92.2} & \textbf{82.5} & \textbf{82.9} & \textbf{74.7} & \textbf{75.7} & \textbf{83.6} & \textbf{91.4} & \textbf{90.6} & \textbf{86.0} & \textbf{74.9} & \textbf{71.1} & \textbf{82.5}
\\
\midrule\midrule
\multicolumn{22}{l}{\textbf{Accuracy only for matched ground truths}}\\
LCR\cite{lcr} & 69.1 & 67.3 & 54.6 & 61.7 & 74.5 & 25.2 & 48.4 & 63.3 & 69.0 & 78.1 & 53.8 & 52.2 & 60.5 & 60.9 & 59.1 & 70.5 & 76.0 & 70.0 & 77.1 & 81.4 & 62.4\\
Singleshot\cite{singleshot} & 81.0 & 65.3 & 64.6 & 63.9 & 75.0 & 30.3 & 65.1 & 61.1 & 64.1 & 83.9 & 72.4 & 69.9 & 71.0 & 72.9 & 71.3 & 83.6 & 79.6 & 73.5 & 78.9 & 90.9 & 70.8\\
LCR++\cite{lcr++} & 88.0 & 73.3 & 67.9 & 74.6 & 81.8 & 50.1 & 60.6 & 60.8 & 78.2 & 89.5 & 70.8 & 74.4 & 72.8 & 64.5 & 74.2 & 84.9 & 85.2 & 78.4 & 75.8 & 74.4 & 74.0\\
Xnect\cite{mehta2019xnect} & 88.4 & 70.4 & 68.3 & 73.6 & 82.4 & 46.4 & 66.1 & 83.4 & 75.1 & 82.4 & 76.5 & 73.0 & 72.4 & 73.8 & 74.0 & 83.6 & 84.3 & 73.9 & 85.7 & 90.6 & 75.8\\
3DMPPE\cite{3dmppe} & \textbf{93.9} & 83.0 & 80.3 & 81.5 & 85.4 & 75.3 & 84.5 & 77.2 & 90.1 & 92.0 & 81.0 & 81.0 & 74.3 & 76.0 & 81.8 & 89.6 & 88.4 & 84.3 & 75.5 & 76.2 & 82.6\\ 
PI-Net (ours) & \textbf{93.9} & \textbf{85.0} & \textbf{81.5} & \textbf{83.0} & \textbf{88.9} & \textbf{75.6} & \textbf{84.7} & \textbf{78.0} & \textbf{90.4} & \textbf{92.2} & \textbf{82.5} & \textbf{82.6} & \textbf{76.0} & \textbf{77.6} & \textbf{83.5} & \textbf{91.5} & \textbf{90.5} & \textbf{85.9} & \textbf{75.7} & \textbf{78.5} & \textbf{83.9}\\
\bottomrule
\end{tabular}}}
\end{center}
\vspace{-0.5cm}
\end{table*}

\subsection{Implementation details}
We use PoseNet of 3DMPPE~\cite{3dmppe} to generate our input 3D human pose. This model is trained on large-scale training data which includes H3.6M single-person 3D dataset~\cite{ionescu2013human3}, MPII\cite{andriluka14cvprmpii} and COCO 2D dataset~\cite{lin2014microsoftcoco}, MuCo multi-person 3D dataset~\cite{singleshot}, and extra synthetic data. 
PI-Net is trained on 33.4k composited MuCo data, which is contained in the training data of the baseline model. This ensures that the improvement of PI-Net comparing with the baseline model is not caused by adding extra training data.

In terms of dimensions, 3DMPPE~\cite{3dmppe} outputs $J= 17$ joints in 3D, the hidden recurrent layers are of dimension $256$, and the Bi-RNN outputs an embedding vector of dimension $E = 512$. We train our PI-net using Adam optimization and the \textit{poly learning rate policy}~\cite{zhao2017pyramid}, with   initial learning rate of 1e-5,  final learning rate of 1e-8, and power of 0.9, for 25 epochs. Batch size is set to 4.

When testing on an image with $n$ instances, we test for $n$ independent  times, each time with a different ordering, and just retain the first person in each case. 

\vspace{0mm}
\section{Experiments}
We next describe the experiment section, which includes a description of the datasets, baselines and evaluation metrics. We then provide a quantitative and qualitative evaluation and comparison to   state-of-the-art approaches. We finalize this section with an exhaustive ablation study of the PI-Net architecture and hyperparameters.

\subsection{Datasets}

\vspace{0mm}
\noindent\textbf{MuCo-3DHP dataset and MuPoTS-3D dataset.}
Most experiments discussed below are performed using these two well-known datasets. They were initially introduced by Mehta et al.~\cite{singleshot} and are typically used as train set and test set respectively, for the task of multi-person 3D human pose estimation.  MuCo-3DHP is a multi-person 3D human pose dataset. Our PI-Net is trained on $33.4k$ MuCo images with $80.7k$ instances, without any other extra data.
MuPoTS-3D test set includes $8320$ images with $23k$ instances in $20$ real scenes ($5$ indoor scenes and $15$ outdoor scenes). Each scene contains from $200$ to $800$ frames extracted from a video, with $2$ or $3$ people performing a certain common activity such as talking, shaking hands or doing sports. These two datasets are annotated using COCO format and provide both 2D image coordinates and 3D camera coordinates for each body joint.

\vspace{1mm}
\noindent\textbf{COCO dataset.}
We also perform qualitative results using the
COCO dataset. This is a large-scale multi-person human pose dataset and, even though it just provides 2D ground truth labels, it depicts challenging scenes with a large number of people performing very diverse  actions. In particular, we use examples from the COCO val2017 subset \cite{maskrcnn}.


\begin{table*}[htbp!]
\caption{PA MPJPE (top) and MPJPE (bottom) comparisons of PI-net with the state-of-the-art method~\cite{3dmppe} used as our baseline on the MuPoTS dataset. The average value indicated image-wise average. Ground truth bounding boxes and roots are used for testing. Lower value means better performance.\vspace{-0.5cm}}
\label{tab:mpjpe(mm)}
\begin{center}
\resizebox{\textwidth}{!}{\setlength{\tabcolsep}{0.7mm}{
\begin{tabular}{l c c c c c c c c c c c c c c c c c c c c c}
\toprule
Sequence & S1 & S2 & S3 & S4 & S5 & S6 & S7 & S8 & S9 & S10 & S11 & S12 & S13 & S14 & S15 & S16 & S17 & S18 & S19 & S20 & AVG\\
\midrule\midrule
\multicolumn{22}{l}{\textbf{PA MPJPE (mm)}} \\\midrule
3DMPPE~\cite{3dmppe} & 67.7 & 102.6 & 82.7 & 82.5 & 79.8 & \textbf{91.1} & 70.8 & 110.1 & 72.8 & 63.5 & 88.6 & 79.6 & 105.1 & 110.5 & 77.5 & 72.2 & 73.3 & 86.8 & 91.9 & 120.0 & 88.4\\
PI-Net (ours) & \textbf{65.8} & \textbf{97.7} & \textbf{82.2} & \textbf{82.4} & \textbf{77.7} & 91.6 & \textbf{68.6} & \textbf{106.3} & \textbf{70.0} & \textbf{60.5} & \textbf{88.0} & \textbf{77.7} & \textbf{102.3} & \textbf{106.6} & \textbf{75.5} & \textbf{70.2} & \textbf{71.5} & \textbf{83.7} & \textbf{88.9} & \textbf{112.6} & \textbf{85.79}\\
\midrule\midrule
\multicolumn{22}{l}{\textbf{MPJPE (mm)}}\\\midrule
3DMPPE~\cite{3dmppe} & 90.9 & 159.3 & 121.8 & 113.5 & 107.8 & \textbf{121.1} & 113.8 & 138.2 & 99.7 & 98.4 & 119.6 & 115.4 & 143.7 & 151.7 & 111.7 & 101.8 & 105.6 & 115.8 & 140.7 & 187.7 & 126.0\\
PI-Net (ours) & \textbf{87.3} & \textbf{151.3} & \textbf{117.1} & \textbf{109.9} & \textbf{103.9} & \textbf{121.1} & \textbf{108.7} & \textbf{133.9} & \textbf{95.8} & \textbf{93.0} & \textbf{117.0} & \textbf{112.2} & \textbf{141.1} & \textbf{146.2} & \textbf{108.0} & \textbf{98.0} & \textbf{102.5} & \textbf{111.8} & \textbf{136.2} & \textbf{178.4} & \textbf{121.7}\\
\bottomrule
\end{tabular}}}
\end{center}
\vspace{-0.5cm}
\end{table*}

\vspace{1cm}
\begin{table*}[htbp!]
\caption{Joint-wise 3DPCK comparison with state-of-the-art methods on the MuPoTS-3D dataset. The first three methods show the reported results in the corresponding paper, the fourth method and our model is tested with ground truth bounding boxes and roots. All ground truths are used for evaluation. Higher value means better performance.\vspace{-0.5cm}}
\label{tab:jointwise pck}
\begin{center}
\setlength{\tabcolsep}{3mm}{
\begin{tabular}{l c c c c c c c c c c c}
\toprule
Method & Hd. & Nck. & Sho. & Elb. & Wri. & Hip & Kn. & Ank. & Avg\\
\midrule\midrule
LCR\cite{lcr} & 49.4 & 67.4 & 57.1 & 51.4 & 41.3 & 84.6 & 56.3 & 36.3 & 53.8\\
single-shot\cite{singleshot} & 62.1 & 81.2 & 77.9 & 57.7 & 47.2 & \textbf{97.3} & 66.3 & 47.6 & 66.0\\
3DMPPE\cite{3dmppe} & \textbf{78.4} & \textbf{91.9} & 83.1 &79.7 & 67.0 & 93.9 & 84.3 & \textbf{75.3} & 81.2 \\
PI-Net (ours) & 78.3	&91.8 & \textbf{87.8} & \textbf{81.9}	& \textbf{68.5}	& 94.2 & \textbf{85.3} & 74.8 & \textbf{82.5}\\
\bottomrule
\end{tabular}}
\end{center}
\vspace{-0.5cm}
\end{table*}
\subsection{Baseline and Evaluation metrics}
\label{section:baseline}
Our pipeline is capable of refining the poses estimated by any   multi-person pose algorithm, independently of the strategy it uses. Given these initially estimated poses we refine them leveraging on the contextual information. In this paper, we use the recent 3DMPPE~\cite{3dmppe} as a baseline and demonstrate both quantitative and qualitative improvements. Note that previous state-of-art works such as PandaNet~\cite{benzine2020pandanet} or SingleShot~\cite{singleshot} do not provide codes either for training or testing, and hence, we could not use them as backbones. 
The baseline~\cite{3dmppe} consists of 3 main steps. Firstly, 2D bounding boxes of humans are detected using Mask-RCNN~\cite{maskrcnn}. For each detection, a deep network refines the coarse root 3D coordinates obtained from camera calibration parameters and, finally, a fully convolutional network~\cite{sun2018integral} predicts root-relative 3D pose. Using the 3D root position, all poses can be represented in a common  camera-coordinates reference.
\newline
We evaluate the performance of all methods by reporting the percentage of keypoints detected by the network that are within 150mm or less from the ground truth labels (3DPCK@150mm). This is the usual evaluation metric on the MuPOTS-3D test set\cite{singleshot,lcr,lcr++,mehta2019xnect,benzine2020pandanet,3dmppe}.
\newline
Notice that the 3DPCK metric depends greatly  on the chosen threshold, for completeness, we also provide MPJPE and PA-MPJPE metrics to evaluate the performances. MPJPE indicates mean-per-joint-position error after root alignment with the ground truth \cite{ionescu2013human3}, and PA-MPJPE denotes MPJPE after Procrustes Alignment\cite{gower1975generalized}. Lower MPJPE and PA-MPJPE indicates better performance.

\subsection{Main results}
\noindent\textbf{Quantitative results on MuPoTS-3D testset.}
We report results of PI-Net on the MuPoTS-3D dataset in Table~\ref{tab:sequencewise pck}, and compare to current state-of-the-art methods. Our results are obtained using the model depicted in Fig.~\ref{fig:pipline}, which uses a bidirectional 3-GRU recurrent layer, followed by a self-attention layer.
We provide results after root alignment with the ground-truth poses, on the two strategies usually used on the MuPoTs datasets. In table\ref{tab:sequencewise pck}, the top-rows \textit{Accuracy for all ground truths} evaluates all annotated persons, and the bottom rows \textit{Accuracy only for matched ground truths} evaluates only predictions matched to annotations by their 2D projections with the 2D ground truths. We got improvements on both of the two strategies. 
PI-Net outperforms all previous models and improves the state-of-the-art by 1.3\% 3DPCK@150mm on average. The improvement is consistent and shows a boost in performance for the majority of actions, setting a new state-of-the-art on the MuPoTS-3D dataset.  Interestingly, we observe that the largest improvements are produced in those actions that require harmony and certain synchronization between people, such as practicing Taekwondo (S2) or playing a ball together (S14).
We use ground truth bounding box and roots to test the baseline, so the root-relative result is comparable with the absolute result here. To avoid the redundancy, we only report root-relative results, which is widely reported in the previous works, for the comparison with the state-of-the-art methods.
\newline
Table~\ref{tab:mpjpe(mm)} shows the comparison of sequence-wise performance using MPJPE with root alignment and PA-MPJPE with further rigid alignment. Testing our model on the MuPoTS test dataset, we reduced the MPJPE error and PA-MPJPE error by 2.6mm and 4.3mm on average, respectively, in comparison with the baseline results~\cite{3dmppe}. Again, results are consistent across different tasks. 
\newline
Table~\ref{tab:jointwise pck} shows a joint-wise comparison with state-of-art methods using 3DPCK@150mm after root alignment with ground truths. While we achieve similar performance with ~\cite{3dmppe} in head, neck and hip, our method consistently outperforms the rest of joints on arms and legs (shoulder, elbow, wrists and knees). Arguably, the joints on the torso have little influence on the interaction between people, which comes mostly through the limbs, for example hands and legs. Hence, it is reasonable that using the context information to refine 3D pose predictions gives the most significant boost in these joints.
\newline
Finally, it is worth pointing out that the results for all previous approaches reported in Tables~\ref{tab:sequencewise pck},~\ref{tab:mpjpe(mm)} and~\ref{tab:jointwise pck} are those of the respective papers. For 3DMPPE~\cite{3dmppe}, however, we tested on ground-truth bounding boxes and roots to reported these results.

\begin{figure*}
\begin{center}
\includegraphics[width=0.8\linewidth]{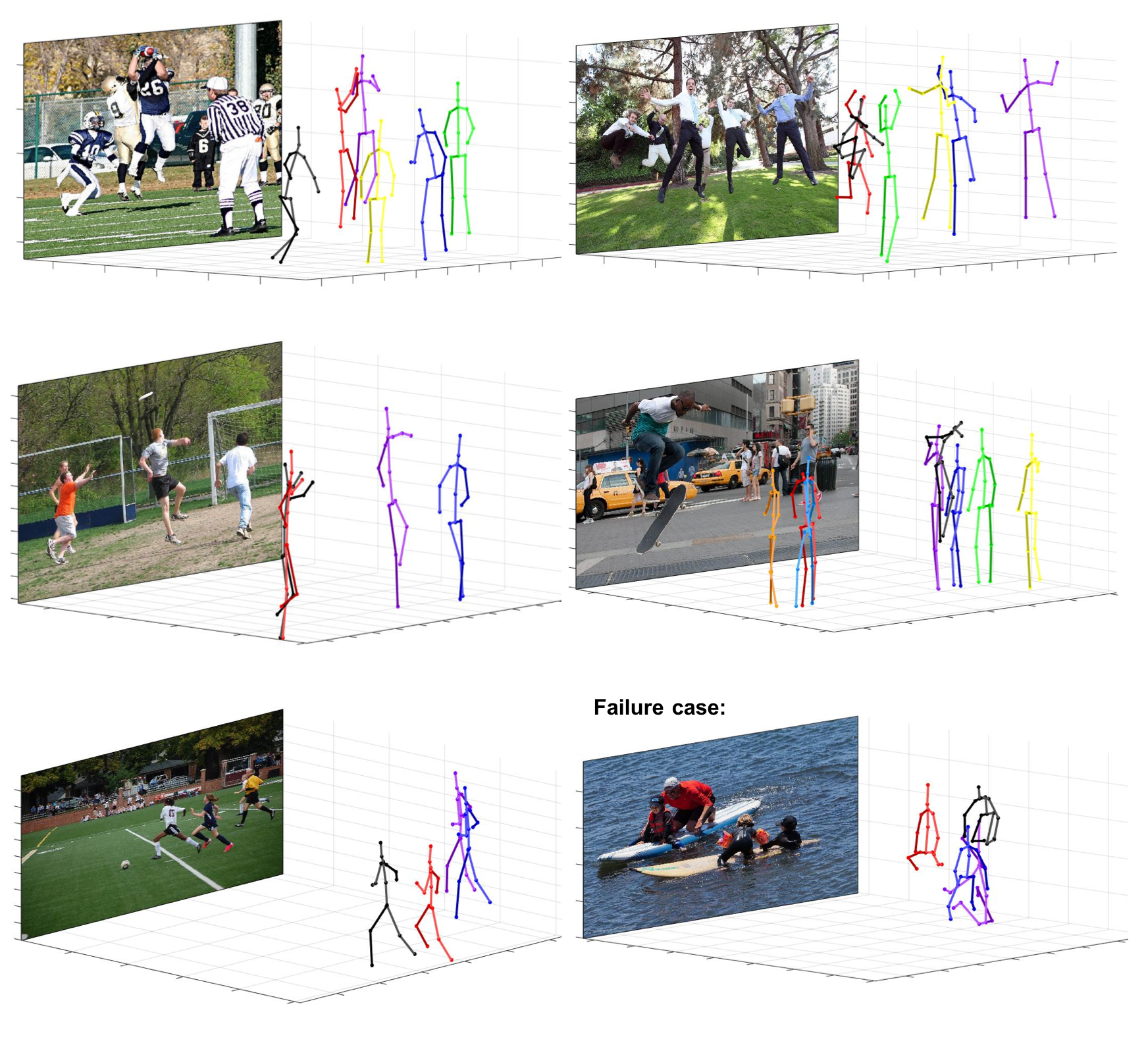}
\vspace{-5mm}
  \caption{{\bf Qualitative results on the COCO dataset}. For each pose, a darker color is used to represent the left side of the person. The bottom-right example corresponds to a failure case, as the `red' and `black' persons should be located in front of the scene, behind the 'blue' and 'purple' persons. 
  This is caused by a misdetection on the root position of the input detected poses provided by the baseline network, while our network designed for refining the poses could not refine this kind of large deviation,
  because this large deviation caused by the baseline network hinder our PI-net from learning the correct context information for correctly interpreting and refining the prediction.}

\label{fig:coco}
\end{center}
\end{figure*}

\vspace{1mm}
\noindent\textbf{Qualitative results on COCO.} Figure~\ref{fig:coco} shows qualitative results on COCO dataset, for which 3D ground truths are not available. We also include (bottom-right) a failure case, caused by a misdetection of the baseline. This is maybe the major  limitation of PI-Net, which is designed to refine poses, but so far, we have not integrated any module to deal with large deviations on the input poses. 

\subsection{Ablation Study}
\label{sec:ablation}
We next provide further analysis of the PI-Net   architectural design and discuss/interpret the predicted adjacency matrix obtained in the self-attention layers.

\begin{table}[t!]
\caption{Comparison of different input orders. \textit{Intuitive} is the one described in Section~\ref{sec:interaction}, from near to far. \textit{Inverse} is the opposite. \textit{Random} means in random order.\vspace{-0.5cm}}
\begin{center}
\begin{tabular}{ c c c c c}
\toprule
Order & PA MPJPE (mm) & MPJPE (mm) \\
\midrule\midrule
Reverse & 86.09 & 122.23  \\
Random & 85.87 & 121.88 \\
Intuitive & \textbf{85.79} & \textbf{121.7} \\
\bottomrule
\end{tabular}
\end{center}
\vspace{-0.5cm}
\label{tab:order}
\end{table}

\vspace{0mm}
\noindent\textbf{Effect of the Input Order.} Table~\ref{tab:order} shows the effect of using different strategies to establish the ordering of the detected people fed into the  Bi-RNN layer. 
We consider three different order: (i) a random ordering, (ii) our approach where we select the person of interest followed by people in order of proximity, and (iii) the inverse approach that person further away is firstly fed into the network. To estimate the distance between people, we compute the distances between the root coordinates of the input people to the target person.

Even though the number of people in images of MuPoTS dataset is relatively small and therefore the results would not differ greatly, the ordering in which every person's information is processed has an effect in the performance of the model. As shown in the Table, the ordering we use provides the best performance and the inverse one results the worst. This demonstrates the importance of taking context into account.

\begin{table}[t!]
\caption{Importance of self-attention and bidirectionality (RNN). PI-Net uses a bidirectional RNN followed by a self-attention layer. We evaluate the impact of each of these choices: w/o Att. when removing attention, w/o Bi. considering standard RNN.\vspace{-0.5cm}}
\begin{center}
\setlength{\tabcolsep}{1mm}{
\begin{tabular}{ l c c c c}
\toprule
Method & PA MPJPE(mm) & MPJPE(mm) \\
\midrule\midrule
PI-Net w/o Att., w/o Bi. & 86.69 & 122.7 \\
PI-Net w/o Bi. & 86.42 & 123.10\\
PI-Net w/o Att. & 85.92 & 122.01 \\ 
PI-Net & \textbf{85.79} &\textbf{121.7} \\ 
\bottomrule
\end{tabular}}
\end{center}
\vspace{-0.5cm}
\label{tab:attention}
\end{table}

\vspace{1mm}
\noindent\textbf{Effect of self-attention and bidirectional RNN.} In Table~\ref{tab:attention} we analyse the effect of using the self-attention layer, which confirms that it helps to boost the performance. We also study the attention weights predicted by the self-attention layer. These weight are, as expected, large at the diagonal, which corresponds to the self-interaction. The larger the distance between two person is, the smaller the weights tend to be. Table~\ref{tab:attention} also compares our approach which employs Bi-RNN with a standard (not bidirectional) RNN. The ablation of the recurrent unit is done later one. Bi-RNN reduces 0.69mm the MPJPE error and 0.77mm the PA-MPJPE error, while the self-attention layers gives an extra improvement of 0.31mm on MPJPE and 0.13mm on PA-MPJPE.

\begin{table}[t!]
\caption{Ablating the unit of the interaction network: None~\cite{3dmppe}, Graph Convolutional Networks (GCN); LSTM and Gated Recursive Units (GRU), with (2,3,4) layers.\vspace{-0.5cm}}
\begin{center}
\setlength{\tabcolsep}{2mm}{
\begin{tabular}{ l c c c c}
\toprule
Interaction & PA MPJPE (mm) & MPJPE (mm) & \# Par.\\
\midrule\midrule
None~\cite{3dmppe} & 88.36 & 126.0 & 133M\\
GCN & 88.67 & 126.3 & 34M\\
\midrule
2 LSTM & 86.45 & 122.5 & 2.78M\\
3 LSTM & 86.17 & 122.3 & 4.36M\\
4 LSTM & 86.32 & 121.7 & 5.93M\\
\midrule
2 GRU & 86.27 & 122.2 & 2.23M \\
\textbf{3 GRU} & \textbf{85.79} & \textbf{121.7} & 3.41M \\
4 GRU & 85.96 & 122.2 & 4.59M\\
\bottomrule
\end{tabular}}
\end{center}
\vspace{-5mm}
\label{tab:lstm}
\end{table}

\vspace{1mm}
\noindent\textbf{Interaction unit.} In Table~\ref{tab:lstm} we report results using alternative units to take the interaction into account. More precisely, Graph Convolution Network (GCN) and LSTM/GRU with different number of layers. For the experiment with GCN, we learned an adjacency matrix for every pair of persons and represented the interaction between them. We considered $4$ GCN layers to obtain the refined poses. We also ablated the recurrent unit: GRU or LSTM~\cite{gers1999lstm}. Even though the MPJPE error of $4$ LSTM layers is similar to that of $3$ GRU layers, we considered the latter because it performs better after rigid alignment, and uses much less parameters which enables it to be trained more efficiently. 

\vspace{-1mm}
\section{Conclusion}

We propose PI-Net, a pose-interacting network that takes initial 3D body poses predicted by any pose estimator, and refine them leveraging on the mutual interaction that occurs in multi-person scenes. We learn such interactions using 3 main building blocks: a bi-directional RNN, a self-attention module, and a MLP. PI-Net is very flexible, lightweight and cost-efficient, and it could improve other approaches for multi-person 3D human pose estimation, establishing the new state-of-the-art. 
This line of work focuses on the interaction between people to improve perception results. In the future, we plan to extend this approach to reason on other contextual information such as objects or structures to better understand human actions and explore different ways to interpret relationships in the scene. Exploiting temporal priors~\cite{hernandeziccv2019,Simo_ijcv2017} and exploring other regression techniques such as robust deep regression~\cite{lathuiliere2018deepgum} or regression adaptation~\cite{fabre2017automatic,hueber2015speaker}, are also other avenues we will explore.

\vspace{-1mm}
\section{Acknowledgement}

  We thank Yuming DU for inspiring discussions and feedback.
  This work has been partially funded by an Amazon Research Award and by the Spanish government under projects HuMoUR TIN2017-90086-R, Maria de Maeztu Seal of Excellence MDM-2016-0656, by the ANR JCJC ML3RI ANR-19-CE33-0008 and the ANR IDEX PIMPE ANR-15-IDEX-02.

{\small
\bibliographystyle{ieee_fullname}
\bibliography{egbib}
}

\end{document}